\documentclass{article}
\usepackage{ijcai20}

\usepackage{times}
\usepackage{soul}
\usepackage{multirow}
\usepackage{url}
\usepackage[hidelinks]{hyperref}
\usepackage[utf8]{inputenc}
\usepackage[small]{caption}
\usepackage{graphicx}
\usepackage{amsmath}
\usepackage{amsthm}
\usepackage{booktabs}
\usepackage{algorithm}
\usepackage{algorithmic}
\usepackage{amsmath}
\usepackage{amssymb,bm}
\usepackage{subfigure}
\usepackage{multirow}
\usepackage{pifont}
\usepackage{dsfont}

\usepackage{xcolor}
\newcommand{\cmark}{\ding{52}} 

\renewcommand{\checkmark}{\cmark}
\newcommand{\nasmall}{non-autoregressive}
\newcommand{\icontext}{time-specific}
\newcommand{\pcontext}{time-agnostic}
\newcommand{\obs}{\text{o}}
\newcommand{\pred}{\text{p}}

\newcommand{\ol}{T_{\obs}}
\newcommand{\pl}{T_{\pred}}

\newcommand{\diag}{\text{diag}}
\newcommand{\name}{NAP}
\newcommand{\etal}{\textit{et al.}}
\newcommand{\ie}{\textit{i.e.}}
\newcommand{\eg}{\textit{e.g.}}

\title{Take a NAP: Non-Autoregressive Prediction for Pedestrian Trajectories}

\author{
Hao Xue
\and
Du. Q. Huynh\and
Mark Reynolds\
\\
The University of Western Australia\\
\emails
hao.xue@research.uwa.edu.au,
du.huynh@uwa.edu.au,
mark.reynolds@uwa.edu.au
}

\begin{document}

\maketitle

\begin{abstract}
Pedestrian trajectory prediction is a challenging task as there are three properties of human movement behaviors which need to be addressed, namely, the social influence from other pedestrians, the scene constraints, and the multimodal (multi-route) nature of predictions.
Although existing methods have explored these key properties, the prediction process of these methods is autoregressive.
This means they can only predict future locations sequentially.
In this paper, we present \name, a \nasmall\ method for trajectory prediction.
Our method comprises specifically designed feature encoders and a latent variable generator to handle the three properties above. 
It also has a time-agnostic context generator and a time-specific context generator for non-autoregressive prediction.
Through extensive experiments that compare \name\ against several recent methods, we show that \name\ has state-of-the-art trajectory prediction performance.

\end{abstract}

\section{Introduction}

Pedestrian trajectory prediction is an important component in a range of applications such as social robots and self-driving vehicles, and plays a key role in understanding human movement behaviors.
This task is not easy due to three key properties in pedestrian trajectory prediction:
(i) \textbf{social interaction}: People are not always walking alone in a public space. Pedestrians often socially interact with others to avoid collisions, walk with friends or keep a certain distance from strangers;
(ii) \textbf{environmental scene constraints}: Besides social interaction, pedestrians' routes also need to obey scene constraints such as obstacles and building layouts; and
(iii) \textbf{multimodal nature of future prediction}: People can follow different routes as long as these routes are both socially and environmentally acceptable. 
For example, a person can choose to turn right or turn left to bypass an obstacle.

Recently, researchers have made progress in incorporating these properties into the trajectory prediction process.
For example, the Social LSTM model~\cite{Alahi_2016_CVPR} is one of the methods that can capture social influence information in a local neighborhood around each pedestrian.
Based on the generative model GAN~\cite{Goodfellow-etal-NIPS14}, the SGAN model proposed by Gupta~\etal~\cite{Gupta_2018_CVPR} can handle multimodality in the prediction process while also capturing the social influence from other pedestrians in the scene.
To deal with the scene constraints, Convolutional Neural Networks (CNNs) are often used to extract scene information in the trajectory prediction network, such as SS-LSTM~\cite{xue2018ss}, SoPhie~\cite{Sadeghian_2019_CVPR}, and Social-BiGAT~\cite{socialbigat_neurips19} .

While other methods like~\cite{su_2017_forecast,vemula2017social,Hasan_2018_CVPR,AAAI18_SAGAIL,xue2019pedestrian,Li_2019_CVPR,zhang2019sr} miss one or two aforementioned key properties, SoPhie~\cite{Sadeghian_2019_CVPR}, Liang~\etal~\cite{Liang_2019_CVPR}, and Social-BiGAT~\cite{socialbigat_neurips19} are three typical papers that have taken all three properties into consideration.
However, these methods predict the future locations recurrently.
There are two main limitations in using autoregression to generate trajectory prediction:
(i)~the autoregressive prediction process works in a recursive manner and so the prediction errors accumulated from previous time steps are passed to the prediction for the next time step;
(ii)~the process cannot be parallelized, \ie, predictions must be generated sequentially, even though one might be interested in generating only the final destination of the pedestrian rather than the entire trajectory. 
 
To overcome the above limitations and inspired by the application of \nasmall\ decoder in other areas such as machine translation~\cite{gu2018nonautoregressive,guo2019non} and time series forecasting~\cite{wen2017multi},
we propose a novel trajectory prediction method that can predict future trajectories non-autoregressively. 
We name our method \textit{NAP} (short for \textit{N}on-\textit{A}utoregressive \textit{P}rediction).
Our research contributions are threefold:
\textbf{(i)}~To the best of our knowledge, we are the first to explore \nasmall\ trajectory prediction. The network architecture of \name\ includes trainable context generators to ensure that context vectors are available for the non-autoregressive decoder to forecast good quality predictions.
The state-of-the-art performance of \name\ is demonstrated through the extensive experiments and ablation study conducted.
\textbf{(ii)}~Both the social and scene influences are handled by \name\ through specially designed feature encoders. The social influence is captured by social graph features propagated through an LSTM; the scene influence is modeled by a CNN.
The effectiveness of these encoders is confirmed from the performance of \name. 
\textbf{(iii)}~Unlike previous work in the literature, \name\ tackles multimodal predictions by training a latent variable generator to learn the latent variables of the sampling distribution for each pedestrian's trajectory. The generator is shown to give \name\ top performance in multimodal predictions.



\newcommand{\bu}{{\bf u}}
\section{Background}
\subsection{Problem Definition}
Pedestrian trajectory prediction is defined as forecasting the future trajectory of the person $i$ given his/her observed trajectory.
We assume that trajectories have already been obtained in the format of time sequences of coordinates (\ie, $\bu^i_t=(x^i_t, y^i_t) \in \mathds{R}^2, \forall i$).
The length of the observed trajectory and predicted trajectory are represented by $\ol$ and $\pl$.
Thus, considering the observed trajectory ${X}^i = \{\bu^i_t \,|\, t= 1,\cdots, \ol \}$,
our target is to generate the prediction $\hat{Y}^i = \{\hat{\bu}^i_t \,|\, t= \ol+1,\cdots, \ol+\pl \}$.

\subsection{Autoregressive and Non-Autoregressive Predictions}
Mathematically, to generate the predicted trajectory $\hat{Y}^i$ from a given observed trajectory $X^i$, the conditional probability $P_\text{AR}(\hat{Y}^i|X^i ; \boldsymbol{\theta})$ with parameter $\boldsymbol{\theta}$ for an autoregressive predictor is defined as:
\begin{equation}
    P_\text{AR}(\hat{Y}^i|X^i ; \boldsymbol{\theta}) = \prod_{t=\ol+1}^{\ol+\pl}P(\hat{\bu}^i_t| \hat{\bu}^i_{\ol:t-1} ,X^i ; \boldsymbol{\theta}), \label{eq:ar}
\end{equation}
where generating the prediction of time step $t$ requires the prediction of all previous time steps in the prediction phase.
This recursive prediction process can not be parallelized.

\begin{figure*}
    \centering
    \includegraphics[width=6in]{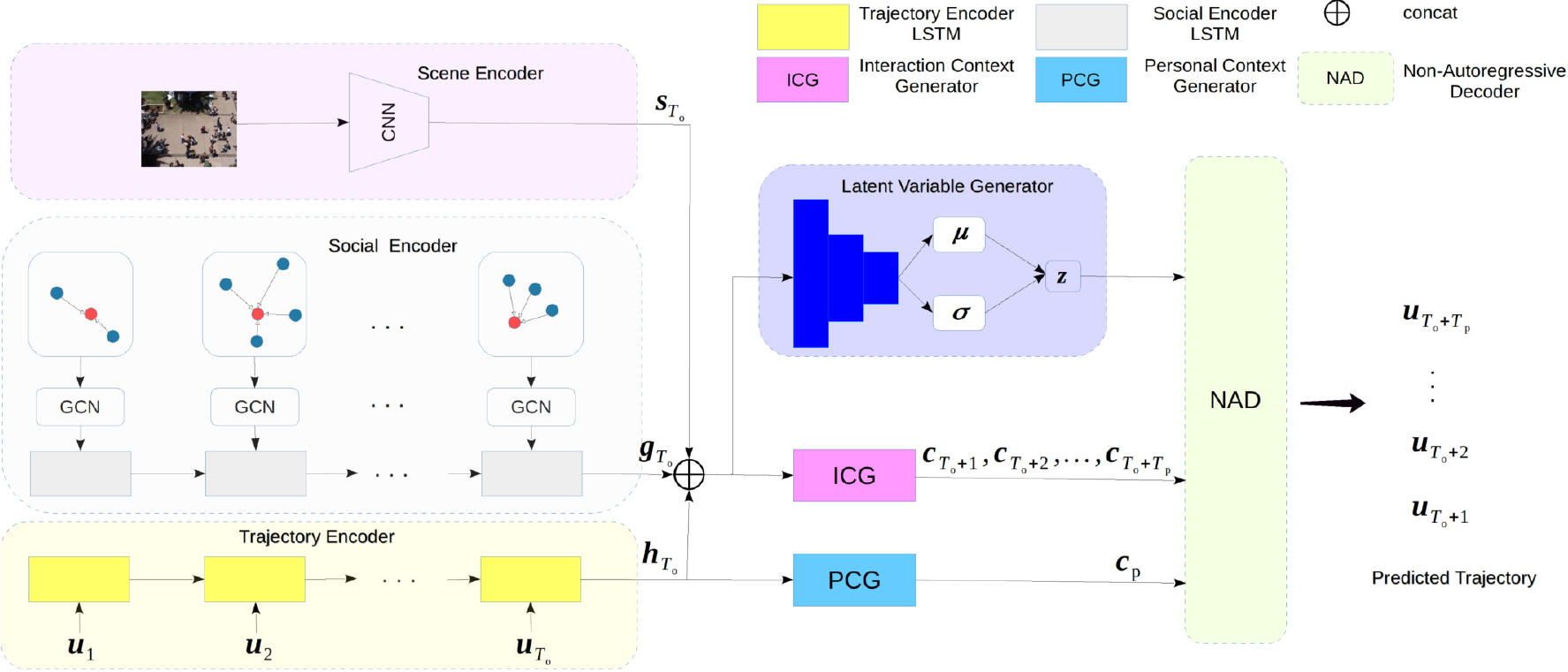}
    \caption{The network architecture of \name.
    There are three encoders to extract features, two context generators to generate context vectors for the non-autoregressive decoder (NAD), and a latent variable generator to handle multimodal predictions.
    The embedding layer and the superscript $i$ (representing the pedestrian index) are not shown to simplify visualization.}
    \label{fig:NAP}
\end{figure*}

Different from the autoregressive prediction process, by treating $\hat{\bu}^i_t$ and $\hat{\bu}^i_{t'}$, for all $t \neq t' \ge \ol$, as independent, the above conditional probability in \nasmall\ predictors becomes:
\begin{equation}
    P_\text{NAR}(\hat{Y}^i|X^i ; \boldsymbol{\theta}) = \prod_{t=\ol+1}^{\ol+\pl}P(\hat{\bu}^i_t|  X^i ; \boldsymbol{\theta}). \label{eq:nar}
\end{equation}
Compared to the autoregression based prediction where the prediction at time step $t$ depends on the information at time step $t-1$, non-autoregression based predictors do not need to generate predictions sequentially. 
However, the removal of this sequential dependency in non-autoregression based methods means that the time awareness ability is compromised in the prediction model, leading to poorer prediction performance.
To compensate for the loss of time awareness ability, we design context generators (detailed in Section~\ref{sec:c_g}) that are trained on the training trajectories. This allows context vectors to be generated from the observed trajectories in the testing stage for the prediction process. As these context vectors can be computed before the start of the prediction phase, predictions at different time steps can be forecast in parallel.

\section{Proposed Method}
\label{sec:3}
Our proposed \name\ comprises four major parts (Fig.~\ref{fig:NAP}):
(i)~feature encoders which are used to encode the input information such as observed trajectories and scene images (Section~\ref{sec:f_e});
(ii)~context generators to yield context vectors for prediction (Section~\ref{sec:c_g});
(iii)~a latent variable generator for multimodality (Section~\ref{sec:z}); and
(iv)~a \nasmall\ decoder for predicting future trajectories (Section~\ref{sec:d}).
Details of these parts are described in the following subsections.

\subsection{Feature Encoders}
\label{sec:f_e}
In \name, there are three feature encoders: 
a trajectory encoder, to learn the representation of the observed history movement of each pedestrian;
a social encoder, to learn the representation of the influence from other pedestrians;
and a scene encoder, to learn the representation of the scene features.

\smallskip
\paragraph{Trajectory Encoder.}

The coordinates of the $i^{\text{th}}$ pedestrian in the observation phase ($t=1, \cdots, \ol $) are firstly embedded into a higher dimensional space through an embedding layer $\phi (\cdot)$.
Then, across different time steps, the embedded features are used as inputs of an LSTM layer (denoted by $\text{LSTM}_{\text{ENC}} (\cdot)$) to get the encoded hidden state $\mathbf{h}^{i}_{t}$ which captures the observed path information.
This trajectory encoding is given by:
\begin{align}
    \label{eq:emb}
    \mathbf{e}^{i}_{t} &= \phi  (x^{i}_{t}, y^{i}_{t}; \mathbf{W}_{\text{EMB}}),\\
    \label{eq:lstm_enc}
    \mathbf{h}^{i}_{t} &= \text{LSTM}_{\text{ENC}}(\mathbf{h}^{i}_{t-1}, \mathbf{e}^{i}_{t};\mathbf{W}_{\text{ENC}}),
\end{align}
where $\mathbf{W}_{\text{EMB}}$ and $\mathbf{W}_{\text{ENC}}$ are trainable weights of the corresponding layers.

\smallskip
\paragraph{Social Encoder.}

At each time step $t$, \name\ captures the social influence on the $i^{\text{th}}$ pedestrian through a graph $\mathcal{G}^i_t=(V^i_t, E^i_t)$.
The $i^{\text{th}}$ pedestrian and all other pedestrians $\mathcal{N}^{(i)}_t$ at the same time step are considered as nodes in the set $V^i_t$.
Edges linking the $i^{\text{th}}$ pedestrian and pedestrians in $\mathcal{N}^{(i)}_t$ form the edge set $E^i_t$.

We then use a graph convolutional network (GCN) to process these graphs.
In the $l^{\text{th}}$ graph convolutional layer, the node feature of pedestrian $i$ is aggregated as follows:
\begin{equation}
    \mathbf{a}^{i,l}_{t} = \text{ReLU}\biggl(\mathbf{b}^l + \frac{1}{ | \mathcal{N}^{(i)}_t |}\sum_{j\in \mathcal{N}_t^{(i)}}\mathbf{W}^l \mathbf{a}^{j,l-1}_{t}\biggr), \label{eq:social_graph}
\end{equation}
where $\mathbf{W}^l$ and $\mathbf{b}^l$ are the weight matrix and bias term.
At the first layer, we initialize the node feature $\mathbf{a}^{i,0}_{t}$ as the location coordinates of the $i^{\text{th}}$ pedestrian, \ie, $\mathbf{a}^{i,0}_{t} = (x_{i}^{t}, y_{i}^{t})$.

The social graph feature $\mathbf{g}^{i}_{t}$ (Eq.~\eqref{eq:social_lstm}) is designed to model the surrounding (or \textit{social}) information of pedestrian $i$ at each time step $t$. To compute this feature, the node features, denoted by $\{\mathbf{a}^{i}_{t} \,|\, t=1,\cdots,\ol\}$, from the final GCN layer across all the time steps in the observation phase are passed through an LSTM layer with trainable weights $\mathbf{W}_{\text{SG}}$, \ie,
\begin{equation}
    \mathbf{g}^{i}_{t} = \text{LSTM}_{\text{SG}}(\mathbf{g}^{i}_{t-1}, \mathbf{a}^{i}_{t};\mathbf{W}_{\text{SG}}). \label{eq:social_lstm}
\end{equation}

\smallskip
\paragraph{Scene Encoder.}

Different from other methods (such as~\cite{xue2018ss,Sadeghian_2019_CVPR}) that process each image frame in the observation phase, the scene encoder of \name\ takes only image $I^i_{\ol}$ as input, since the scene encoder focuses on the static information like scene layouts and obstacles. Not only does this save the computation time, but it also supplies the most up-to-date and sufficient scene context before prediction kicks in at $t=\ol+1$. 
We use a CNN to model the scene feature $\mathbf{s}^{i}_{\ol}$ as follows:
\begin{equation}
    \mathbf{s}^i_{\ol} = \text{CNN}(I^i_{\ol}; \mathbf{W}_{\text{CNN}}).
    \label{eq:cnn}
\end{equation}
We take the merit of DeepLabv3+~\cite{Chen_2018_ECCV}, a state-of-the-art semantic segmentation architecture, and set the initial value of $\mathbf{W}_{\text{CNN}}$ to the weight matrix from DeepLabv3+ that is pre-trained on the Cityscapes dataset~\cite{Cordts2016Cityscapes}.

\subsection{Context Generators}
\label{sec:c_g}
The role of the context generators is to aggregate the outputs of the feature encoders for the downstream decoder for trajectory forecasting. We use two context generators in \name: (i)~a \textit{personal context generator} that is \textit{\pcontext}, as its input is the hidden state $\mathbf{h}^{i}_{t}$ computed from the $i^{\text{th}}$ pedestrian's own trajectory only; (ii)~an \textit{interaction context generator} that is \textit{\icontext} as its input includes both social graph and scene interaction features also.

\paragraph{Personal Context Generator (PCG).}
We use a Multi-Layer Perceptron (MLP) to model this context generator. The output context vector $\mathbf{c}^{i}_{\text{p}}$ is computed as
\begin{equation}
    \mathbf{c}^{i}_{\text{p}} = \text{MLP}_{\text{A}}(\mathbf{h}^{i}_{\ol};\mathbf{W}_{\text{A}}), \label{eq:personal}
\end{equation}
where $\mathbf{W}_{\text{A}}$ is the corresponding weight matrix. 
The context $\mathbf{c}^{i}_{\text{p}}$ captures the ``personal'' cues such as the pedestrian's preferable walking speed and direction in the observation phase, oblivious of his/her surrounding.
This context is \textit{\pcontext} because, without considering the social and scene influences, such personal cues can remain the same for the entire trajectory, \eg, the pedestrian can continue to walk in a straight line or at a fast pace with no penalty when bumping into obstacles or other pedestrians since both the social graph and scene features are not present in the equation. 

\smallskip
\paragraph{Interaction Context Generator (ICG).}
This context generator incorporates both the social graph and scene features. These two types of influences allow the context generator to be time-specific, \eg, while a pedestrian can walk at a fast pace in the initial part of his/her trajectory, he/she would need to slow down or detour at a later part of the trajectory in order to avoid other pedestrians or scene obstacles. Similar to PCG, we use an MLP to model ICG but its input, being the concatenation of $\mathbf{h}^{i}_{\ol}$, $\mathbf{g}^{i}_{\ol}$, and $\mathbf{s}^i_{\ol}$, contains richer information. The output of ICG comprises different context vectors for different time steps in the prediction phase, as given below:
\begin{equation}
    (\mathbf{c}^{i}_{\ol+1}, \mathbf{c}^{i}_{\ol+2}, \cdots, \mathbf{c}^{i}_{\ol+\pl}) = \text{MLP}_{\text{B}}(\mathbf{h}^{i}_{\ol}\oplus \mathbf{g}^{i}_{\ol}\oplus \mathbf{s}^i_{\ol};\mathbf{W}_{\text{B}} ),
    \label{eq:interaction}
\end{equation}
where $\mathbf{W}_{\text{B}}$ is the corresponding weight matrix.

\subsection{Latent Variable Generator}
\label{sec:z}

For multimodal prediction, \name\ is designed to generate multiple trajectories through the latent variables $\boldsymbol{\mu}$ and $\boldsymbol{\sigma}$ (see Fig.~\ref{fig:NAP}).
Although several existing trajectory prediction methods such as~\cite{Lee_2017_CVPR,Gupta_2018_CVPR,Li_2019_CVPR,huang2019stgat} also use latent variables to handle multimodality, the latent variables in these methods are either directly sampled from the normal distribution or a multivariate normal distribution conditioned on the observed trajectories.
To make our latent variables more aware of the social and scene cues,
we design \name\ to learn the parameters ($\boldsymbol{\mu}_i$ and $\boldsymbol{\sigma}_i$) of the sampling distribution from the observed trajectories, the social influence, and the scene influence features.
To this end, the concatenated feature $\mathbf{h}^{i}_{\ol} \oplus \mathbf{g}^{i}_{\ol} \oplus \mathbf{s}^i_{\ol}$ is passed to two different MLPs (Eqs.~\eqref{eq:mu}-\eqref{eq:sigma}) to yield the mean vector $\bm{\mu}_i$ and  variance $\bm{\sigma}_i$ and finally $\mathbf{z}_i$ for the downstream non-autoregressive decoder:
\begin{align}
    \label{eq:mu}
    \bm{\mu}_i &=  \text{MLP}_{\mu}(\mathbf{h}^{i}_{\ol}\oplus \mathbf{g}^{i}_{\ol}\oplus \mathbf{s}^i_{\ol};\mathbf{W}_{\mu} ), \\
    \label{eq:sigma}
    \bm{\sigma}_i &= \text{MLP}_{\sigma}(\mathbf{h}^{i}_{\ol}\oplus \mathbf{g}^{i}_{\ol}\oplus \mathbf{s}^i_{\ol};\mathbf{W}_{\sigma} ),\\
    \label{eq:sample}
    \mathbf{z}_i &\sim \mathcal{N}(\bm{\mu}_i, \diag(\bm{\sigma}_i^2)),
\end{align}
where
$\mathbf{W}_{\mu} $ and $\mathbf{W}_{\sigma}$ are trainable weights of $\text{MLP}_{\mu}(\cdot)$ and $\text{MLP}_{\sigma}(\cdot)$.
The reparameterization trick~\cite{kingma2013auto} is applied to sample the latent variable $\mathbf{z}_i$.

\subsection{Non-Autoregressive Decoder (NAD)}
\label{sec:d}

\begin{figure}[t]
    \centering
    \includegraphics[width=1.5in]{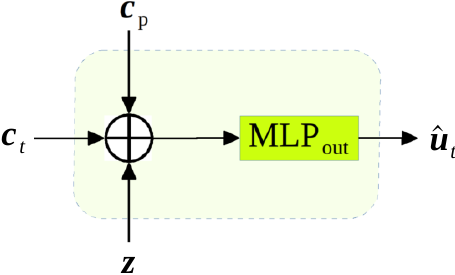}
    \caption{Details of the non-autoregressive decoder (NAD) of \name.}
    \label{fig:nad}
\end{figure}

In the work of~\cite{gu2018nonautoregressive}, the authors introduce in their model a component that enhances the inputs passed to the decoder for their machine translation problem. The idea is to help the model learn the internal dependencies (which are absent in their non-autoregressive translator) within a sentence. In the work of~\cite{guo2019non}, the authors use a positional module to improve the decoder's ability to perform local reordering. The context generators in \name\ play a similar role as these two approaches. In the testing stage, the trained PCG and ICG are able to generate context vectors for new (unseen) observed trajectories to help the decoder improve its time awareness for trajectory prediction. While the ICG, which generates \icontext\ contexts $\{\mathbf{c}_{t}\,|\,\ol\!+\!1 \leqslant t \leqslant \ol\!+\!\pl\}$, is obviously more important than the PCG, the \pcontext\ PCG is also needed so as to help the NAD tie with the specific trajectory being considered.


To make multimodal predictions at time step $t$, the NAD therefore takes the concatenation of the two contexts $\mathbf{c}^{i}_{\text{p}}$ and $\mathbf{c}^{i}_{t}$ and the latent variable $\mathbf{z}_i$ as inputs, modeled by the MLP below (see Fig.~\ref{fig:nad}):
\begin{equation}
    (\hat{x}_t^i, \hat{y}_t^i) = \text{MLP}_{\text{out}}(\mathbf{c}^{i}_{t}\oplus \mathbf{c}^{i}_{\text{p}}\oplus \mathbf{z}_i;\mathbf{W}_{\text{out}} ),
    \label{eq:nad}
\end{equation}
where $\text{MLP}_{\text{out}}(\cdot)$ is the MLP used for predicting the location coordinates.
Its parameter $\mathbf{W}_{\text{out}}$ is shared across all the time steps in the prediction phase. 
Note that the input passed to the NAD in Fig.~\ref{fig:nad} is different for each time step $t$ in the prediction phase as $\mathbf{c}^i_t$ depends on $t$.
We can consider that the contexts $\{\mathbf{c}^i_t\}$ function like the hidden states in the decoder of an LSTM except that they are not recursively defined.




\subsection{Implementation Details}
The embedding layer $\phi$ in Eq.~\eqref{eq:emb} is modeled as a single-layer perceptron that outputs 32-dimensional embedded vectors for the input location coordinates.
The dimensions of the hidden states of the LSTM layers for both the Trajectory and Social Encoders are 32.
The GCN in the Social Encoder is a single graph convolutional layer (\ie, $l=1$ in Eq.~\eqref{eq:social_graph}).
For the ICG, $\text{MLP}_{\text{B}}$ is a three-layer MLP with ReLU activations.
All the other MLPs used in Eqs.~\eqref{eq:personal}, \eqref{eq:mu}, \eqref{eq:sigma}, and~\eqref{eq:nad} are single-layer MLPs.
Except for in Section~\ref{sec:pl} where we explore the prediction performance for different prediction lengths, the observed length of input trajectories is 8 time steps ($\ol = 8$) and the prediction length is 12 time steps ($\pl=12$) for all other experiments.

We implemented \name\ and its variants (Section~\ref{sec:ablation}) using the PyTorch framework in Python.
The Adam optimizer was used to train our models with the learning rate set to 0.001 and the mini-batch size to 128.

\section{Experiments}
\label{sec:4}
\subsection{Datasets and Metrics}

We use the popular ETH~\cite{pellegrini2009you} and UCY~\cite{lerner2007crowds} datasets, which, altogether, include 5 scenes: ETH, HOTEL, UNIV, ZARA1, and ZARA2.
Similar to~\cite{Sadeghian_2019_CVPR,zhang2019sr}, we normalize each pedestrian's coordinates and augment the training data by rotating trajectories.
As raw scene images are used as inputs for extracting scene features, we also rotate the input images when input trajectories are rotated.
Same as previous work in the literature~\cite{Alahi_2016_CVPR,Gupta_2018_CVPR,Sadeghian_2019_CVPR,huang2019stgat}, the leave-one-out
strategy is adopted for training and testing.
All methods are evaluated based on two standard metrics: the Average Displacement Error (ADE) and the Final Displacement Error (FDE).
Smaller errors indicate better prediction performance.

\begin{table*}
\renewcommand{\dagger}{\star}
\centering
\footnotesize
\addtolength{\tabcolsep}{-0.2ex}
\begin{tabular}{|lc||c|c|c|c|c||c|}
\hline
\multirow{2}{*}{Method} &
\multirow{2}{*}{\#}&
\multicolumn{6}{c|}{ETH \& UCY scenes} \\ \cline{3-8} 
 &  & ETH & HOTEL & UNIV & ZARA1 & ZARA2 & Average \\ \hline
Social-LSTM~\cite{Alahi_2016_CVPR}$^\ddagger$  &  & 1.09 / 2.35 & 0.79 / 1.76 & 0.67 / 1.40 & 0.47 / 1.00 & 0.56 / 1.17 & 0.72 / 1.54 \\ 
SGAN 1V-1~\cite{Gupta_2018_CVPR}$^\dagger$  &  & 1.13 / 2.21 & 1.01 / 2.18 & 0.60 / 1.28 & 0.42 / 0.91 & 0.52 / 1.11 & 0.74 / 1.54 \\
MX-LSTM~\cite{Hasan_2018_CVPR}$^\dagger$  &  & -- & -- & \textbf{0.49} / \textbf{1.12} & 0.59 / 1.31 & 0.35 / 0.79 & -- \\ 
Nikhil \& Morris~\cite{nikhil2018convolutional}$^\dagger$  & & 1.04 / 2.07 & 0.59 / 1.17 & 0.57 / 1.21 & 0.43 / 0.90 & 0.34 / 0.75 & 0.59 / 1.22 \\ 
Liang~\etal~\cite{Liang_2019_CVPR}$^\dagger$ &  & 0.88 / 1.98 & 0.36 / 0.74 & 0.62 / 1.32 & 0.42 / 0.90 & 0.34 / 0.75 & 0.52 / 1.14 \\ 
MATF~\cite{Zhao_2019_CVPR}$^\dagger$  &  & 1.33 / 2.49 & 0.51 / 0.95 & 0.56 / 1.19 & 0.44 / 0.93 & 0.34 / 0.73 & 0.64 / 1.26 \\ 
SR-LSTM~\cite{zhang2019sr}$^\dagger$  &  & 0.63 / 1.25 & 0.37 / 0.74 & {0.51} / {1.10} & \textbf{0.41} / 0.90 & 0.32 / 0.70 & \textbf{0.45} / 0.94 \\ 
STGAT 1V-1~\cite{huang2019stgat}$^\dagger$  &  & 0.88 / 1.66 & 0.56 / 1.15 & 0.52 / 1.13 & \textbf{0.41} / 0.91 & \textbf{0.31} / \textbf{0.68} & 0.54 / 1.11 \\ \hline
\name\ (ours) &  & \textbf{0.59} / \textbf{1.13} & \textbf{0.30} / \textbf{0.51} & 0.59 / 1.23 & \textbf{0.41} / \textbf{0.86}  & 0.36 / 0.72  & \textbf{0.45} / \textbf{0.89} \\ \hline \hline
SGAN 20V-20~\cite{Gupta_2018_CVPR}$^\dagger$  & \checkmark & 0.81 / 1.52 & 0.72 / 1.61 & 0.60 / 1.26 & 0.34 / 0.69 & 0.42 / 0.84 & 0.58 / 1.18 \\ 
SoPhie~\cite{Sadeghian_2019_CVPR}$^\dagger$  & \checkmark & 0.70 / 1.43 & 0.76 / 1.67 & 0.54 / 1.24 & 0.30 / 0.63 & 0.38 / 0.78 & 0.54 / 1.15 \\ 
Liang~\etal~\cite{Liang_2019_CVPR}$^\dagger$  & \checkmark & 0.73 / 1.65 & 0.30 / 0.59 & 0.60 / 1.27 & 0.38 / 0.81 & 0.31 / 0.68 & 0.46 / 1.00 \\ 
MATF GAN~\cite{Zhao_2019_CVPR}$^\dagger$  & \checkmark & 1.01 / 1.75 & 0.43 / 0.80 & \textbf{0.44} / \textbf{0.91} & 0.26 / 0.45 & 0.26 / 0.57 & 0.48 / 0.90 \\ 
IDL~\cite{Li_2019_CVPR}$^\dagger$  & \checkmark & 0.59 / 1.30 & 0.46 / 0.83 & 0.51 / 1.27 & \textbf{0.22} / \textbf{0.49} & \textbf{0.23} / \textbf{0.55} & 0.40 / 0.89 \\ 
STGAT 20V-20~\cite{huang2019stgat}$^\dagger$  & \checkmark & 0.65 / 1.12 & 0.35 / 0.66 & 0.52 / 1.10 & 0.34 / 0.69 & 0.29 / 0.60 & 0.43 / 0.83 \\ 
Social-BiGAT~\cite{socialbigat_neurips19}$^\dagger$  & \checkmark & 0.69 / 1.29 & 0.49 / 1.01 & 0.55 / 1.32 & 0.30 / 0.62 & 0.36 / 0.75 & 0.48 / 1.00 \\ \hline
\name\ (ours) & \checkmark & \textbf{0.53} / \textbf{1.08} & \textbf{0.26} / \textbf{0.46} & 0.58 / 1.22 & 0.30 / 0.65 & 0.28 / 0.60 & \textbf{0.39} / \textbf{0.80} \\ \hline
\end{tabular}
\vspace{-1.5ex}
\caption{ The ADEs / FDEs (in meters) of various methods. The settings are: $\ol=8$ and $\pl=12$. The results with a $\dagger$ are taken from the authors' papers. The result with $\ddagger$ is taken from [Gupta~\etal, 2018].}
\label{tab:all_results}
\end{table*}

\subsection{Comparison with Other Methods}

We compare \name\ against the following state-of-the-art trajectory prediction methods:
Social-LSTM~\cite{Alahi_2016_CVPR},
SGAN~\cite{Gupta_2018_CVPR},
MX-LSTM~\cite{Hasan_2018_CVPR},
Nikhil \& Morris~\cite{nikhil2018convolutional},
Liang~\etal~\cite{Liang_2019_CVPR},
MATF~\cite{Zhao_2019_CVPR},
SR-LSTM~\cite{zhang2019sr},
SoPhie~\cite{Sadeghian_2019_CVPR},
IDL~\cite{Li_2019_CVPR},
STGAT~\cite{huang2019stgat}, and
Social-BiGAT~\cite{socialbigat_neurips19}.

In Table~\ref{tab:all_results}, all the compared methods are put into two groups depending on whether they generate only one prediction (top half of the table) or multiple predictions (bottom half and indicated by a tick under the \# column) for each input observed trajectory.
The multimodal predictions being considered is 20. The reported ADEs and FDEs are computed from the best predictions out of the 20.
Five methods report both single and multimodal prediction results and so they appear in both halves of the table: SGAN, MATF, STGAT, Liang~\etal, and \name.

Our proposed method is able to achieve results on par with the state-of-the-art methods for the single prediction setting.
Specifically, \name\ has the same smallest average ADE (0.45m) as SR-LSTM 
while outperforming all methods on the average FDE (0.89m).
In addition to \name, SR-LSTM, MX-LSTM, and STGAT 1V-1 also have the best performance on one or more scenes.
In the lower half of Table~\ref{tab:all_results}, results of multimodal predictions are given and compared.
On average, our \name\ achieves the smallest ADE of 0.39m and the smallest FDE of 0.80m.
For each scene, the best performers that achieve the smallest ADE/FDE in the lower half of the table include \name, IDL, MATF, and GAN.
Taken together, these results demonstrate the efficacy of our proposed method in both single and multimodal prediction settings.

\subsection{Ablation Study}
\label{sec:ablation}
To explore the effectiveness of different contexts working together in our proposed method, we consider four variants of \name\ listed below:
\begin{itemize}
\setlength{\itemsep}{0ex}
    \item \name-P: This variant only uses the Personal Context Generator (\pcontext\ context, the light blue PCG box in Fig.~\ref{fig:NAP}), 
    \ie, the interaction context $\mathbf{c}^{i}_{t}$ in Eq.~\eqref{eq:nad} is removed.
    \item \name-ISS: In contrast to \name-P,  \name-ISS disables the personal context and forecasts predictions based on the \icontext\ interaction context (the pink box in Fig.~\ref{fig:NAP}). 
    The personal context $\mathbf{c}^{i}_{\text{p}}$ in Eq.~\eqref{eq:nad} is removed.
    The rest of \name-ISS is the same as \name.
    \item \name-ISg: In order to further investigate the impact of removing the scene influence, we drop the scene feature $\mathbf{s}^i_{\ol}$ from the Interaction Context Generator in Eq.~\eqref{eq:interaction} to form this variant. That is, the interaction context $\mathbf{c}^{i}_{t}$ in \name-ISS is computed using both the social graph and scene features, whereas the $\mathbf{c}^{i}_{t}$ in \name-ISg is computed using the social graph feature only.
    \item \name-ISc: Similar to \name-ISg, this variant is designed to investigate the impact of removing the social influence.
    We drop the social graph feature $\mathbf{g}^{i}_{\ol}$ but keep the scene feature $\mathbf{s}^i_{\ol}$ in Eq.~\eqref{eq:interaction} so the context $\mathbf{c}^{i}_{t}$ is computed from the scene feature only.
\end{itemize}

\begin{table}
\centering
\footnotesize
\begin{tabular}{|l||c|c|c|c|}
\hline
 & \name-P & \name-ISS & \name-ISg & \name-ISc \\ \hline
ETH & 0.87 / 1.63  &  0.66 / 1.22 & 0.69 / 1.31 & 0.74 / 1.52 \\
HOTEL & 0.43 / 0.77 & 0.34 / 0.61& 0.37 / 0.70 & 0.38 / 0.73   \\
UNIV & 0.71 / 1.42 & 0.68 / 1.37& 0.68 / 1.35 & 0.70 / 1.39   \\
ZARA1 & 0.46 / 0.95 &  0.45 / 0.94& 0.47 / 0.96 & 0.45 / 0.95  \\
ZARA2 & 0.44 / 0.88 &  0.42 / 0.83& 0.44 / 0.84 & 0.44 / 0.86  \\\hline
Average & 0.58 / 1.13 & 0.51 / 0.99& 0.53 / 1.03 & 0.54 / 1.09   \\\hline
\end{tabular}
\vspace{-1.5ex}
\caption{The ADEs / FDEs (in meters) of the four variants for single predictions in the ablation study.}
\label{tab:ablation}
\end{table}

In our ablation study, we compare only the single prediction performance (see Table~\ref{tab:ablation}) of these four variants, \ie,  
the latent variable $\mathbf{z}_i$ for multimodality is removed from Eq.~\eqref{eq:nad} in the experiments.
In general, \name-P, which uses only the personal context (\pcontext), has a poorer performance than the other three variants. This is not unexpected as, without the time-specific context, the 
NAD is not able to forecast good predictions for different time steps in the prediction phase.
Comparing the three interaction context based variants against each other, it is not surprising to see that \mbox{\name-ISS} outperforms the other two variants due to the presence of both the social graph and scene features.
As for \mbox{\name-ISg} against \name-ISc, we observe that \name-ISg slightly outperforms \name-ISc.
This demonstrates that the social influence is more important than the scene influence. However, it should be noted that the five scenes in the ETH/UCY datasets do not have many obstacles scattered in the pedestrian pathways. The slightly better performance of \mbox{\name-ISg} confirms that there are more social (pedestrian) interactions than scene interactions in these datasets.

Comparing the results from the four variants in Table~\ref{tab:ablation} and from \name\ in Table~\ref{tab:all_results}, we observe that \name\ outperforms all the four variants. Our ablation study justifies the need for all the contexts to be present in \name.

\subsection{Different Prediction Lengths}
\label{sec:pl}

\begin{table}[]
\footnotesize
\centering
\addtolength{\tabcolsep}{-0.22ex}
\begin{tabular}{|l||c|c|c|}
\hline
 & $\pl=8$ & $\pl=12$ & Increment \\ \hline
Social-LSTM & 0.45 / 0.91 & 0.72 / 1.54 & 60.00\% / 69.23\% \\ 
SGAN 1V-1 & 0.49 / 1.00 & 0.74 / 1.54 & 51.02\% / 54.00\% \\ 
STGAT 1V-1 & 0.39 / 0.81 & 0.54 / 1.11 & 38.46\% / 37.03\% \\ \hline
NAP (ours) & 0.35 / 0.67 & 0.45 / 0.89 & \textbf{28.57\%} / \textbf{32.84\%} \\ \hline\hline
SGAN 20V-20$^*$ & 0.39 / 0.78 & 0.58 / 1.18 & 48.72\% / 51.28\% \\ 
STGAT 20V-20$^*$ & 0.31 / 0.62 & 0.43 / 0.83 & 38.71\% / 33.87\% \\ \hline
NAP (ours)$^*$ & 0.31 / 0.61 & 0.39 / 0.80 & \textbf{25.81\%} / \textbf{31.15\%} \\ \hline
\end{tabular}
\vspace{-1.5ex}
\caption{The ADEs / FDEs (in meters) of various methods for different prediction lengths. A method with $^*$ indicates that it generates 20 predictions for each input observed trajectory.}
\label{tab:tp}
\end{table}

In addition to the prediction length setting ($\pl=12$ frames, corresponding to 4.8 seconds) used in Tables~\ref{tab:all_results} \&~\ref{tab:ablation} and similar to previous work such as SGAN~\cite{Gupta_2018_CVPR} and STGAT~\cite{huang2019stgat},
we conduct experiments for the prediction length $\pl=8$ frames (or 3.2 seconds) to further evaluate the performance of \name. 
Table~\ref{tab:tp} shows the average ADE/FDE results for this prediction length setting. The figures under the `$\pl=12$' column are copied from the \textit{Average} column of Table~\ref{tab:all_results}. Each error increment (last column) due to the increase of $\pl$ is calculated as: $(e_{\text{p}12} - e_{\text{p}8}) / e_{\text{p}8} \times 100\%$, where $e_{\text{p}12}$ and $e_{\text{p}8}$ are errors (ADEs or FDEs) for $\pl\!=\!12$ and $\pl\!=\!8$ of the same method.

As expected, all methods shown in Table~\ref{tab:tp} have better performance for the shorter prediction length.
In the top half of the table, when generating a predicted trajectory for each input, the error increments of Social-LSTM and \mbox{SGAN 1V-1} are over 50\%.
Compared to these two methods, STGAT 1V-1 has smaller error increments for both ADE and FDE.
For the multimodal predictions (bottom half of the table), STGAT 20V-20 again
outperforms SGAN 20V-20. 

We observe from Table~\ref{tab:tp} that \name\ consistently outperforms all other methods for both prediction length settings and for both single and multimodal predictions. Furthermore, \name\ also has the smallest error increments for both ADE and FDE when $\pl$ increases.
This demonstrates that \name\ is more robust in generating long trajectories. The reason is due to the non-autoregressive nature of the decoder, which not only allows the location coordinates at different time steps to be independently forecast but also helps minimize the accumulation of prediction errors when the prediction length increases.


\subsection{Qualitative Results}

\begin{figure}[t]
  \centering
  \subfigure[ETH]{
  \includegraphics[trim=100 125.0 80 25, clip,width=3.7cm]{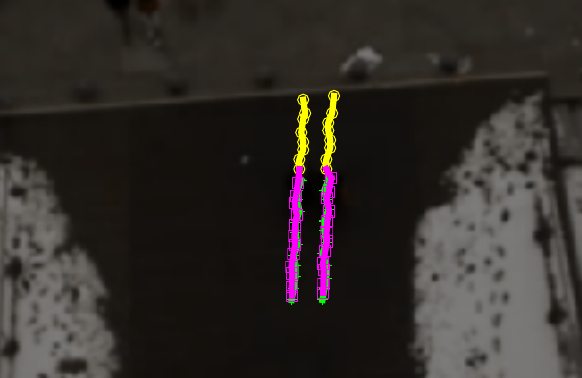}}
\subfigure[ZARA2]{
  \includegraphics[trim=35 40.0 15 40, clip,width=3.7cm]{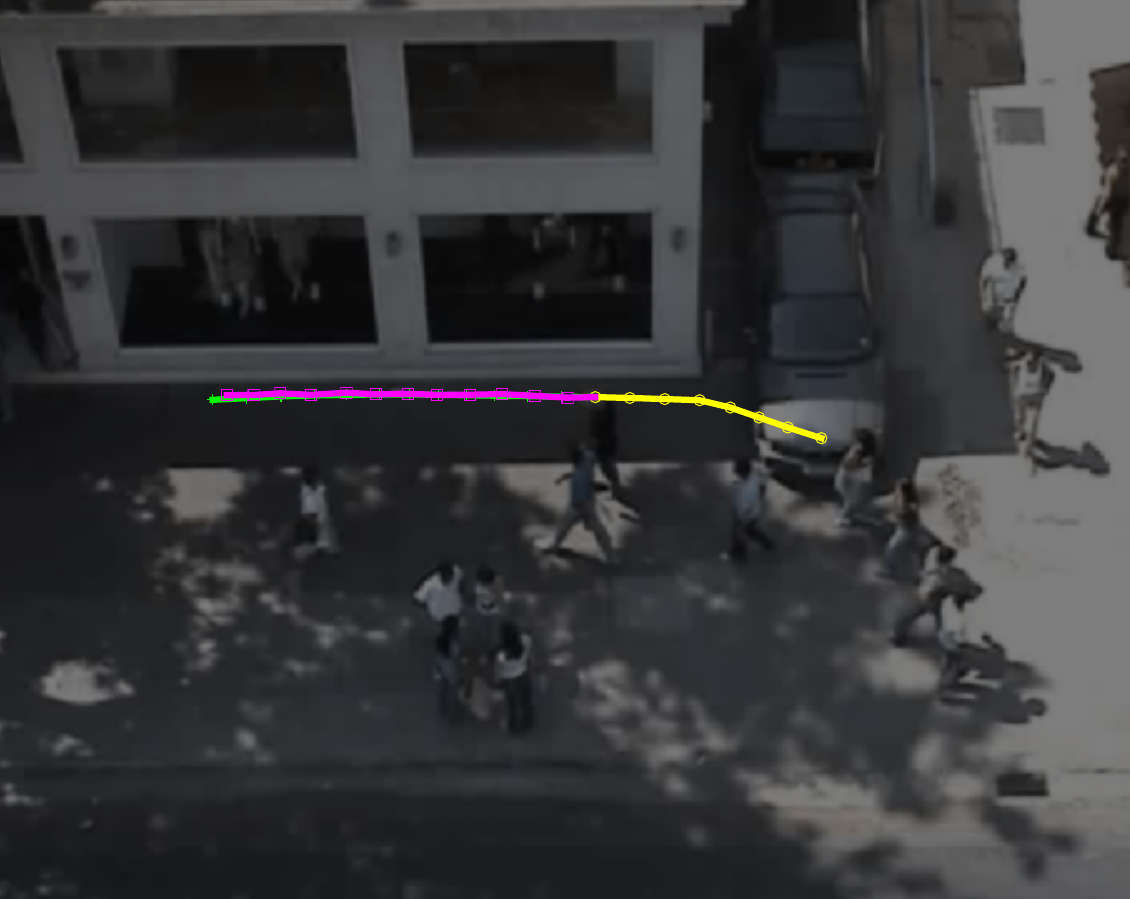}} \\ [-0.2cm]
  \subfigure[ZARA1]{
  \includegraphics[trim=35 40.0 15 40, clip,width=3.7cm]{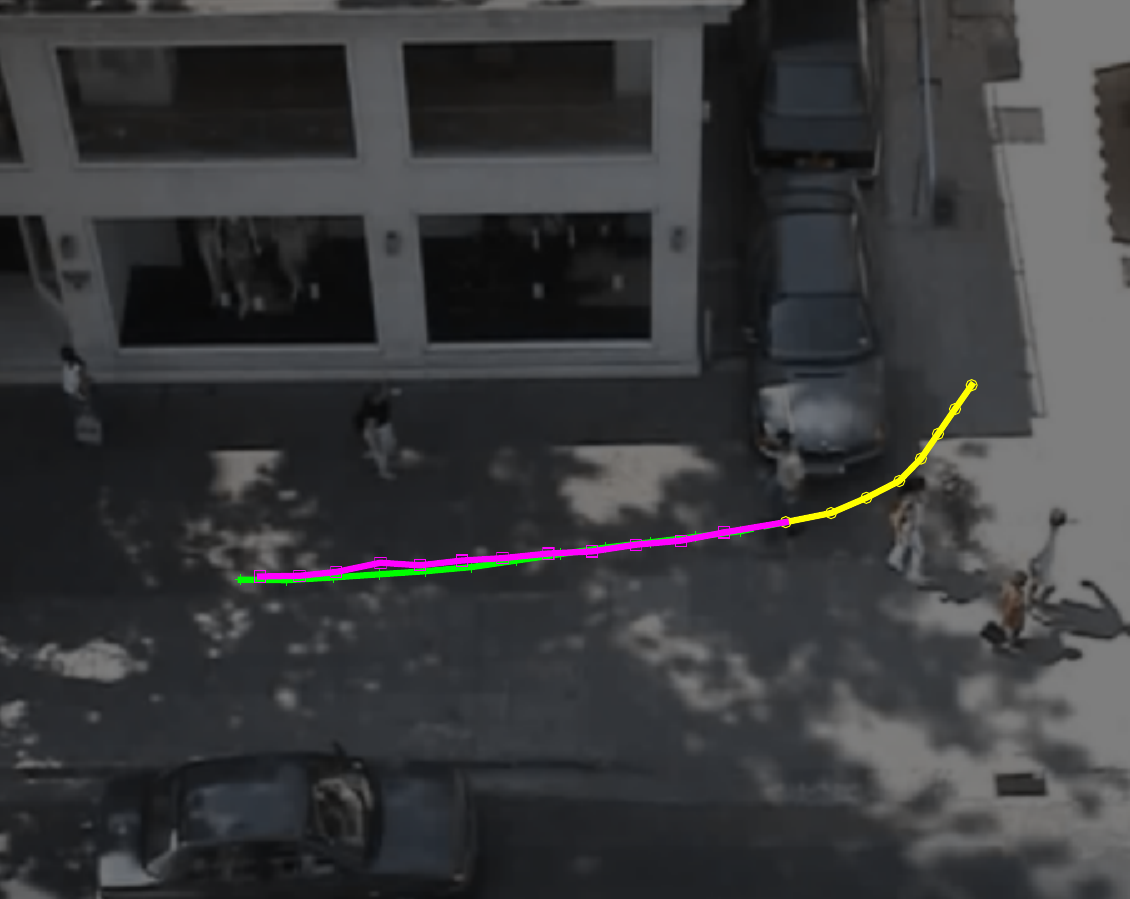}}
  \subfigure[ZARA1]{
  \includegraphics[trim=15 40.0 35 40, clip,width=3.7cm]{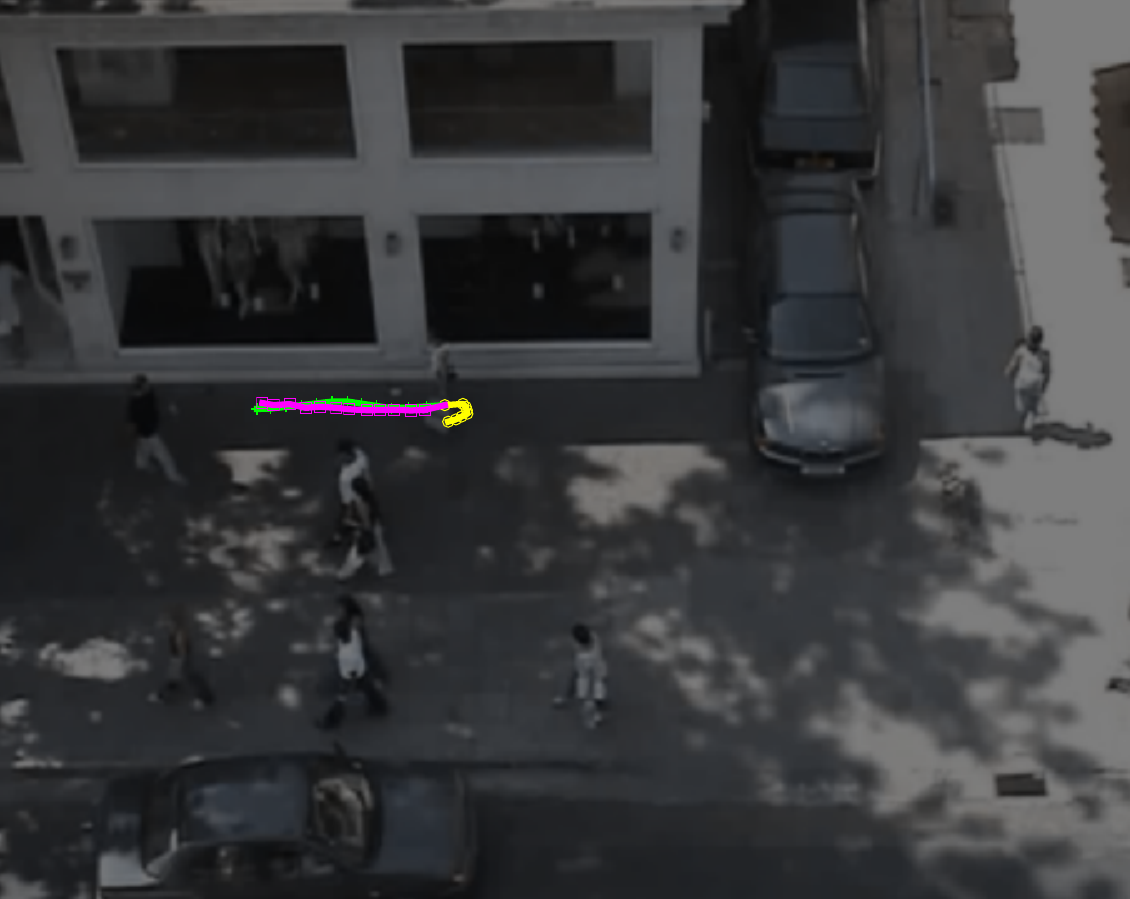}}
  \vspace{-0.4cm}
\caption{Examples of predicted trajectories (shown in pink) generated by \name. The observed trajectories and ground truth trajectories are shown in yellow and green (figure best viewed in color).}
\label{fig:vis}
\end{figure}

Figure~\ref{fig:vis} illustrates some prediction examples generated by \name.
The observed and ground truth trajectories are shown in yellow and green;
the best trajectory of the 20 predictions of each pedestrian is shown in pink.
For better visualization, the video frames have been darkened and blurred.
These examples cover different movement behaviors of pedestrians.
For example, Fig.~\ref{fig:vis}(a) shows two simple straight path scenarios, Fig.~\ref{fig:vis}(b) and (c) show a gentle turning scenario, and
Fig.~\ref{fig:vis}(d) shows a more difficult scenario in which an abrupt turning occurs near the end of the observation phase. 
Although the predicted trajectory (in pink) in Fig.~\ref{fig:vis}(d) does not perfectly overlap with the ground truth trajectory, \name\ is still able to correctly predict the trajectory from the late turning cue.

\begin{figure}[t]
  \centering
  \subfigure[HOTEL]{
  \includegraphics[trim=35 70.0 15 10, clip,width=3.7cm]{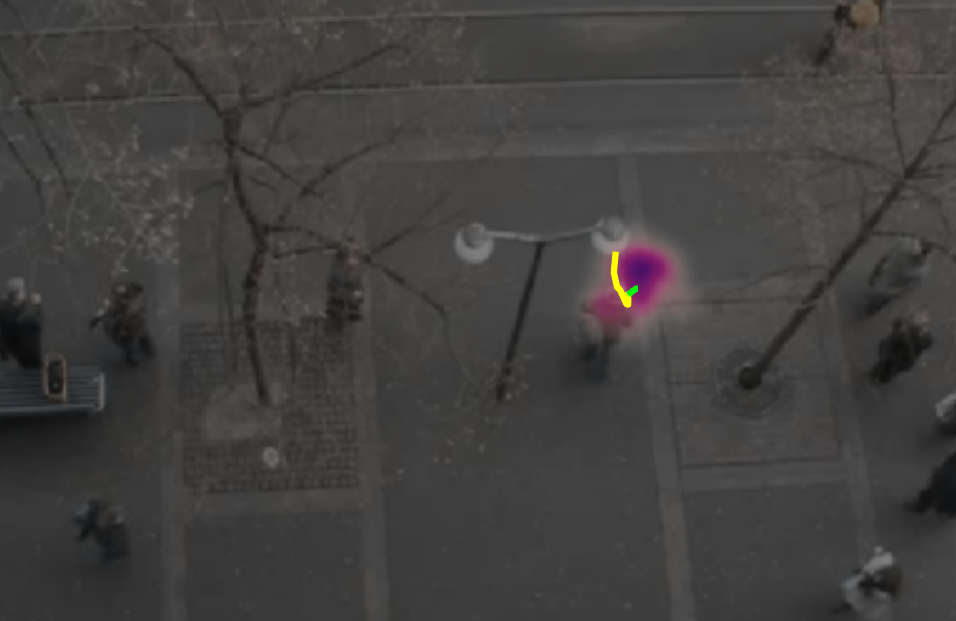}}
\subfigure[HOTEL]{
  \includegraphics[trim=35 40.0 15 40, clip,width=3.7cm]{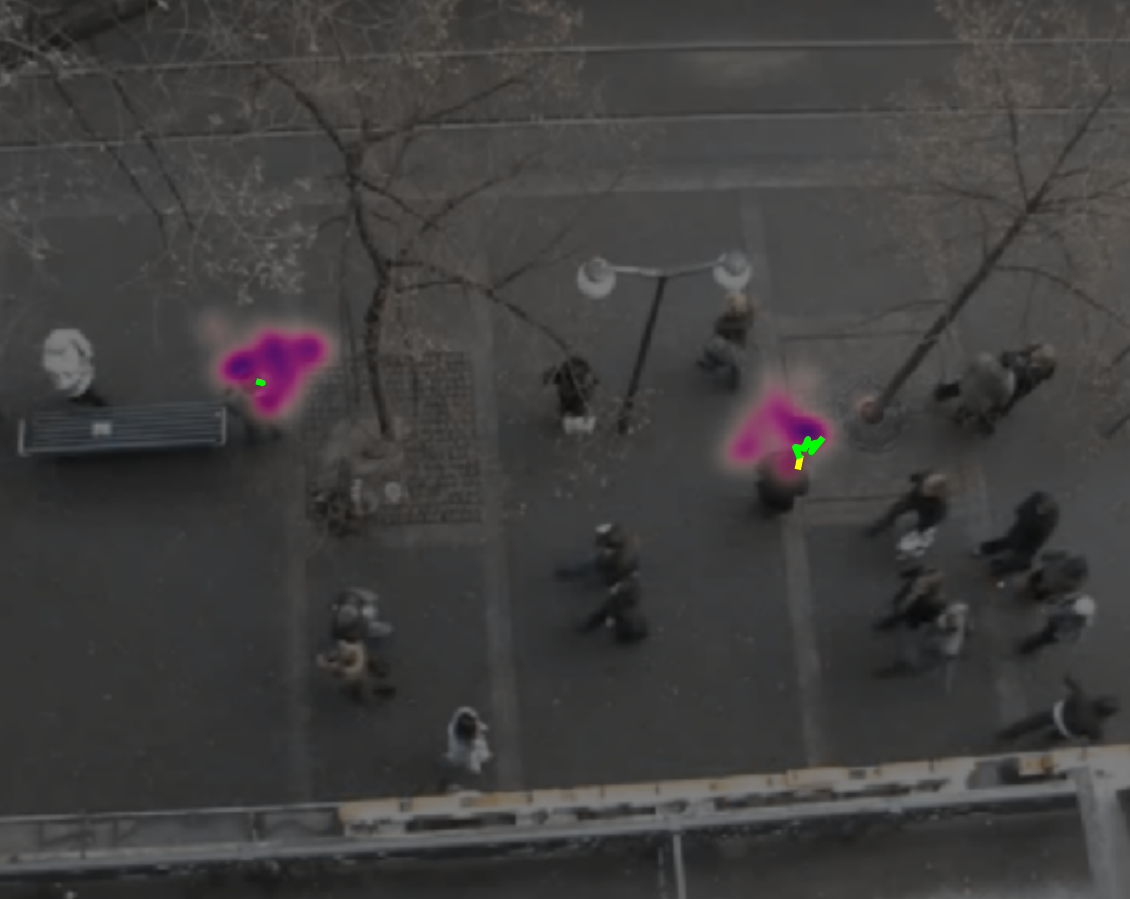}}
  \vspace{-0.4cm}
\caption{Examples of multiple predicted trajectories shown as heatmaps.
The observed trajectories and ground truth trajectories are shown in yellow and green 
(figure best viewed in color).}
\label{fig:heatmap}

\end{figure}

Figure~\ref{fig:heatmap} shows two more difficult scenarios where all the 20 predicted trajectories are displayed as a heatmap around each pedestrian.
For the pedestrian in Fig.~\ref{fig:heatmap}(a) and the right pedestrian in Fig.~\ref{fig:heatmap}(b), each made an abrupt turn at almost the last frame of the observation phase. However, \name\ is still able to give good predicted trajectories, as all the plausible paths (including the ground truth trajectories (green)) are well covered by the heatmaps.
The left pedestrian in Fig.~\ref{fig:heatmap}(b) is a stopping scenario. After stopping, the pedestrian can remain still or resume walking in any direction. The generated heatmap shows a good coverage of possible paths; however, it has a small dent in the bottom left hand region due to the presence of a bench there, showing that the pedestrian must bypass the obstacle. This example shows the importance of including scene influence in the method.


\section{Conclusion}
\label{sec:5}
We have presented a novel method called \name\ which can handle both social influence and scene influence in the pedestrian trajectory prediction process. \name\ captures these influences using the trainable feature encoders in the network. In addition, \name\ handles multimodal predictions via a latent variable generator which models the sampling distribution that describes the multiple plausible paths of each pedestrian. Unlike existing trajectory prediction methods, the decoder of \name\ is non-autoregressive. \name\ is therefore able to forecast predictions for different time steps simultaneously or to forecast only for those time steps that are of interest. From our extensive experiments and ablation study, the context encoders used in \name\ have been demonstrated to be effective. Not only does \name\ achieve state-of-the-art performance, it also has smaller error increments when the prediction length increases.


\bibliographystyle{plain}
\bibliography{nap_hao_2020}

\end{document}